\RequirePackage{tikz}
\documentclass[referee, pdflatex, sn-apa]{sn-jnl}

\usepackage{subcaption}
\usetikzlibrary{positioning,angles,quotes,arrows,shapes}
\usetikzlibrary{bayesnet}
\tikzset{>=latex}

\usepackage{adjustbox}

\DeclareMathOperator*{\argmax}{arg\,max}
\newcommand{\E}{\mathbb{E}}

\DeclareMathOperator*{\dirichlet}{Dirichlet}
\DeclareMathOperator*{\categorical}{Categorical}

\DeclareMathOperator*{\TS}{TruncatedStudentT}

\DeclareMathOperator*{\ub}{ub}

\jyear{2021}%

\begin{document}

\title[POMDP inference and robust solution via deep reinforcement learning]{POMDP inference and robust solution via deep reinforcement learning: An application to railway optimal maintenance}


\author*[1]{\fnm{Giacomo} \sur{Arcieri}}\email{giacomo.arcieri@ibk.baug.ethz.ch}

\author[1]{\fnm{Cyprien} \sur{Hoelzl}}\email{hoelzl@ibk.baug.ethz.ch}

\author[2]{\fnm{Oliver} \sur{Schwery}}\email{oliver.schwery@sbb.ch}

\author[3]{\fnm{Daniel} \sur{Straub}}\email{straub@tum.de}

\author[4]{\fnm{Konstantinos G.} \sur{Papakonstantinou}}\email{kpapakon@psu.edu}

\author[1]{\fnm{Eleni} \sur{Chatzi}}\email{chatzi@ibk.baug.ethz.ch}

\affil*[1]{\orgdiv{Institute of Structural Engineering}, \orgname{ETH Z{\"u}rich}, \orgaddress{\city{Z{\"u}rich}, \postcode{8093},\country{Switzerland}}}

\affil[2]{ \orgname{Swiss Federal Railways SBB}, \orgaddress{\city{Bern}, \postcode{3000}, \country{Switzerland}}}

\affil[3]{\orgdiv{Engineering Risk Analysis Group}, \orgname{Technical University of Munich}, \orgaddress{\city{Munich}, \postcode{80333}, \country{Germany}}}

\affil[4]{\orgdiv{Dept. of Civil and Environmental Engineering}, \orgname{Pennsylvania State Univ.}, \orgaddress{\street{University Park},\postcode{16802}, \state{PA}, \country{USA}}}

\abstract{Partially Observable Markov Decision Processes (POMDPs) can model complex sequential decision-making problems under stochastic and uncertain environments. A main reason hindering their broad adoption in real-world applications is the lack of availability of a suitable POMDP model or a simulator thereof. Available solution algorithms, such as Reinforcement Learning (RL), require the knowledge of the transition dynamics and the observation generating process, which are often unknown and non-trivial to infer. In this work, we propose a combined framework for inference and robust solution of POMDPs via deep RL. First, all transition and observation model parameters are jointly inferred via Markov Chain Monte Carlo sampling of a hidden Markov model, which is conditioned on actions, in order to recover full posterior distributions from the available data. The POMDP with uncertain parameters is then solved via deep RL techniques with the parameter distributions incorporated into the solution via domain randomization, in order to develop solutions that are robust to model uncertainty. As a further contribution, we compare the use of transformers and long short-term memory networks, which constitute model-free RL solutions, with a model-based/model-free hybrid approach. We apply these methods to the real-world problem of optimal maintenance planning for railway assets.}


\maketitle

\section{Introduction}\label{sec:intro}
Partially Observable Markov Decision Processes (POMDPs) offer a mathematically sound framework to model and solve complex sequential decision-making problems \citep{cassandra1998survey}. POMDPs account for the uncertainty associated with observations in order to derive optimal policies, namely a sequence of optimal decisions that minimize/maximize the total costs/rewards over a prescribed time horizon, under stochastic and uncertain environments. Stochasticity can indeed be incorporated both in the evolution of the hidden states over time, i.e., the transition dynamics, and in the process that generates the observations, which reflect only a partial and/or noisy information of the actual states. 

POMDPs  form a potent mathematical framework to model optimal maintenance planning for deteriorating engineered systems \citep{papakonstantinou2014planning2}. In such problems, a perfect information of the system's condition (state) is generally not available or feasible to acquire, due to the problem's scale, inherent noise of sensing instruments, and associated costs limitations. 
By using sensors and inferred associated condition indicators, Structural Health Monitoring (SHM) tools, as described by \citet{farrar2012structural, straub2017value, andriotis2021value}, can provide estimates of the structural state. However, the provided observations are often incomplete and susceptible to noise, which limits their ability to accurately determine the true state of the system. Consequently, decision-making must occur in the face of irreducible uncertainty. Within a POMDP scheme, the decision maker (or agent) receives an observation from an SHM system, using it to form a belief about the current state of the system. Based on this belief, the agent takes an action, which will impact the future condition of the system. The POMDP objective is to find the optimal sequence of maintenance actions that minimizes the expected total costs over the operating life-cycle. A list of applications of POMDP modeling to optimal maintenance can be found in \citep{madanat1994optimal, ellis1995inspection, durango2002optimal, memarzadeh2015optimal, schobi2016maintenance, papakonstantinou2018pomdp, kivancc2022maintenance}.

POMDP solutions assume knowledge of the transition dynamics and the observation generating process. This implies strict prior assumptions on the POMDP model parameters that govern the deterioration, the effects of maintenance actions, and the relation of observations to latent states and variables. When a POMDP model is available, the solution can be computed via Dynamic Programming (DP) \citep{bertsekas2012dynamic} and approximate methods \citep{papakonstantinou2014planning} with optimality convergence guarantees, when the complexity of the problem is not prohibitive, or via Reinforcement Learning (RL) schemes \citep{sutton2018reinforcement} through samples and trial and error learning. While RL methods can relax some assumptions on the POMDP knowledge, a simulator that can reliably describe the POMDP model is still necessary for inference and testing purposes, particularly for engineering problems and in infrastructure asset management applications.

However, a full POMDP model of the problem is rarely available in real-world applications, and its inference can be quite challenging. The availability of such a model is a key issue that prevents wide adoption of the POMDP framework and its solution methods (e.g., reinforcement learning) for real-world applications. Available literature on the theme of maintenance planning is focused on developing RL methods to solve complex POMDP problems, as pioneered by the work of \citet{andriotis2019managing, andriotis2021deep}, while assuming knowledge of the POMDP transition and observation models, i.e., by for example assuming that the POMDP inference has already been carried out. Only few papers deal with the POMDP inference, which poses a challenge in itself, while best practices are not generally available. \citet{papakonstantinou2014planning2, song2022value, wari2023discrete} propose methods to estimate the state transition probability matrix for deterioration processes, but without demonstrating inference on the transition matrices associated with maintenance actions. \citet{guo2022predictive} propose methods to estimate both the transition and the observation models, but do not consider model uncertainty and the implementation examples do not involve real-world data but only simulated ones.

In \citet{arcieri2022bridging}, we tackle this key inference issue by proposing a framework to jointly infer all transition and observation model parameters entirely from available real-world data, via Markov Chain Monte Carlo (MCMC) sampling of a Hidden Markov Model (HMM), which is conditioned on actions. The framework, which is relatively easy to implement and can be tailored to the problem at hand, estimates full posterior distributions of POMDP model parameters. By considering these distributions in the POMDP evaluation, optimal policies that are robust with respect to POMDP model uncertainties are obtained.

In this work, we combine the POMDP inference with a deep RL solution. Most previous works on deep RL methods focus on fully observable problems, with RL solutions for POMDPs having received notably lower attention. Partial observability is usually overcome with deep learning architectures that are able to infer hidden states through memory and a history of past observations. \citet{schmidhuber1990reinforcement} is one of first works that applied Recurrent Neural Networks (RNNs) for RL problems. Subsequently, Long Short-Term Memory (LSTM) networks have become the standard to handle partial observability \citep{dung2008reinforcement, zhu2017improving, meng2021memory}. Recent works propose to replace LSTM architectures with Transformers (GTrXL) \citep{parisotto2020stabilizing}. A third modeling option, which constitutes a hybrid approach between a DP and a RL solution, exploits the POMDP model to compute beliefs via Bayes' theorem, which are then fed to the deep RL algorithm as inputs to classical feed-forward Neural Networks (NNs) \citep{andriotis2019managing, andriotis2021deep, morato2023inference}. Namely, the POMDP problem is converted into the belief-MDP \citep{papakonstantinou2014planning, andriotis2021value} and then solved with deep RL techniques. We compare these three available solution methods and propose a joint framework of inference and robust solution of POMDPs based on deep RL techniques, by combining MCMC inference with domain randomization of the RL environment in order to incorporate model uncertainty into the policy learning.

We showcase the applicability of these methods and of the proposed framework on a real-world problem of optimal maintenance planning for railway infrastructure. The problem, modelled as a POMDP, is based on on-board railway monitoring data, namely the so-called ``fractal values'' condition indicator, computed from field measurements and provided by our SBB (the Swiss Federal Railways) partners.

The remainder of this paper is organized as follows. Section \ref{sec:preliminaries} provides the necessary background on POMDPs. 
Section \ref{sec:data} describes the considered maintenance planning problem of railway assets and the monitoring data. Section \ref{sec:inference} describes the POMDP inference and its implementation to the problem here considered. Section \ref{sec:ComparisonSolution} evaluates the three available modeling options of deep RL solutions for POMDPs, namely LSTM, GTrXL, and the belief-input case. Section \ref{sec:DomainRandomization} proposes our joint framework of POMDP inference and robust solution via deep RL and domain randomization. Finally, Section \ref{sec:conc} concludes with a highlight and a discussion of the contributions, and outlines possible future work.

\section{Preliminaries}\label{sec:preliminaries}

\subsection{Partially Observable Markov Decision Processes}
A POMDP can be considered as a generalization of a Markov Decision Process (MDP) for modelling sequential decision-making problems within a stochastic control setting, with uncertainty incorporated into the observations. A POMDP is defined by the tuple $\langle S,A,Z,R,T,O,b_0,H,\gamma\rangle$, where:
\begin{itemize}
    \item $S$ is the finite set of hidden states that the environment can assume. 
    \item $A$ is the finite set of available actions. 
    \item $Z$ is the set of possible observations, generated by the hidden states and executed actions, which provide partial and/or noisy information about the actual state of the system. 
    \item $R:S \times A  \rightarrow\mathbb{R}$ is the reward function that assigns the reward $r_t=R(s_t, a_t)$ for assuming an action $a_t$ at state $s_t$. 
    \item $T:S \times S \times A  \rightarrow [0,1]$ is the transition dynamics model that describes the probability $p(s_{t+1}\mid s_t,a_t)$ to transition to state $s_{t+1}$ if action $a_t$ is taken at state $s_t$.
    \item $O:S \times A \times Z  \rightarrow \mathbb{R}$ is the observation generating process that defines the emission probability $p(z_t\mid s_t,a_{t-1}, z_{t-1})$, namely the likelihood to observe $z_t$ if the system is at state $s_t$ and action $a_{t-1}$ was taken.
    \item $b_0$ is the initial belief on the system's state $s_0$.
    \item $H$ is the considered horizon of the problem, which can be finite or infinite. 
    \item $\gamma$ is the discount factor that discounts future rewards to obtain the present value. 
\end{itemize}

In the POMDP setting, the agent takes a decision based on a formulated belief over the system's state. Such a belief is defined as a probability distribution over $S$, which maps the discrete finite set of states into a continuous $\mid S\mid-1$ dimensional simplex \citep{papakonstantinou2014planning}. It is a sufficient statistics over the complete history of actions and observations. Solving a POMDP is thus equivalent to solving a continuous state MDP defined over the belief space, termed the belief-MDP \citep{papakonstantinou2014planning, andriotis2021value}. The belief over the system's state is updated according to Bayes' rule every time the agent receives a new observation:
\begin{equation}\label{eq:belief}
    b(s_{t+1})=\frac{p(z_{t+1}\mid s_{t+1},a_t)}{p(z_{t+1}\mid\mathbf{b},a_t)}\sum_{s_t\in S}p(s_{t+1}\mid s_t,a_t)b(s_t)
\end{equation}
where the denominator is the normalizing factor:
\begin{equation}
    p(z_{t+1}\mid \mathbf{b},a_t)=\sum_{s_{t+1}\in S}p(z_{t+1}\mid s_{t+1},a_t)\sum_{s_t\in S}p(s_{t+1}\mid s_t,a_t)b(s_t)
\end{equation}

The objective of the POMDP is to determine the optimal policy $\pi^*$, which maps beliefs to actions, that maximizes the expected sum of rewards:
\begin{equation}\label{eq:obj}
    \pi^*=\argmax_\pi\E\left[\sum_{t=0}^H\gamma^tr_t\right]
\end{equation}
where $r_t=R(s_t,\pi(b_t))$. Algorithms based on DP \citep{bertsekas2012dynamic} can be used to compute the optimal policy. These algorithms rely on two key functions: the value function $V^\pi$, which calculates the expected sum of rewards for a policy $\pi$ starting from a given state until the end of the prescribed horizon, and the Q-value function $Q^{\pi}$ \citep{sutton2018reinforcement}, which estimates the expected value for assuming action $a_t$ in state $s_t$, and then following policy $\pi$.

\begin{figure}[htb]
\begin{tikzpicture}[auto,node distance=8mm,>=latex,font=\small]
\tikzset{
    >=stealth',
    node distance=1.5cm,
    state/.style={minimum size=50pt,font=\small,circle,draw},
    dots/.style={state,draw=none},
    edge/.style={->},
    trans/.style={font=\footnotesize,above=2mm},
    reflexive/.style={out=120,in=60,looseness=5,relative},
    squared/.style={rectangle, draw=black, fill=white, thick, minimum size=8mm},
    round/.style={circle, draw=black, fill=white, thick, minimum size=8mm},
    round_hidden/.style={circle, draw=black, fill=gray!25, thick, minimum size=8mm},
    decision/.style={diamond, draw=black, fill=white, thick, minimum size=8mm},
    minimum size=8mm,inner sep=0pt
  }
    
    \node[round] (b0) {$b_t$};
    \node[round_hidden] (o0) [above=15mm of b0] {$z_t$};
    \node[round] (s0) [above=15mm of o0] {$s_t$};
    \node [dots]  (d0)  [left=20mm of b0] {};
    \node [dots]  (ds0)  [left=20mm of s0] {};
    \node [dots]  (do0)  [left=20mm of o0] {$\cdots$};
    \node[squared] (a0) [below right=5mm and 5mm of s0] {$a_t$};
    \node[decision] (r0) [above right=10mm and 5mm of s0] {$r_t$};
    \node[round,right=20mm of b0] (b1) {$b_{t+1}$};
    \node[round_hidden] (o1) [above=15mm of b1] {$z_{t+1}$};
    \node[round] (s1) [above=15mm of o1] {$s_{t+1}$};
    \node[squared] (a1) [below right=5mm and 5mm of s1] {$a_{t+1}$};
    \node[decision] (r1) [above right=10mm and 5mm of s1] {$r_{t+1}$};
    \node [dots]  (d1)  [right=20mm of b1] {};
    \node [dots]  (ds1)  [right=20mm of s1] {};
    \node [dots]  (do1)  [right=20mm of o1] {$\cdots$};
    
    \draw[->] (d0)--(b0);
    \draw[->] (ds0)to node[above right=-1.5mm and 0.mm]{ $T$}(s0);
    \draw[->] (b0)--(b1);
    \draw[->] (s0)to node[above right=-1.5mm and 0.mm]{ $T$}(s1);
    \draw[->] (s0)to node[below left=0mm and -1mm]{ $O$}(o0);
    \draw[->] (do0)--(o0);
    \draw[->] (o0)--(b0);
    \draw[->] (o0)--(o1);
    \draw[->] (b0)to node[above right=-5mm and -3mm]{ $\pi$}(a0);
     \draw[->] (s0)--(r0);
    \draw[->] (a0)--(b1);
    \draw[->] (a0)--(s1);
    \draw[->] (a0)--(o1);
     \draw[->] (a0)to node[above left=3.5mm and -1mm]{ $R$}(r0);
    \draw[->] (b1)--(d1);
    \draw[->] (s1)to node[above right=-1.5mm and 0.mm]{ $T$}(ds1);
    \draw[->] (s1)to node[below left=0mm and -1mm]{ $O$}(o1);
    \draw[->] (o1)--(b1);
    \draw[->] (o1)--(do1);
    \draw[->] (b1)to node[above right=-5mm and -3mm]{ $\pi$}(a1);
    \draw[->] (s1)--(r1);
    \draw[->] (a1)to node[above left=3.5mm and -1mm]{ $R$}(r1);
    \draw[->] (a1)--(d1);
    \draw[->] (a1)--(ds1);
    \draw[->] (a1)--(do1);
\end{tikzpicture}
\caption{Probabilistic graphical model of a POMDP.}
\label{fig:pomdp}
\end{figure}
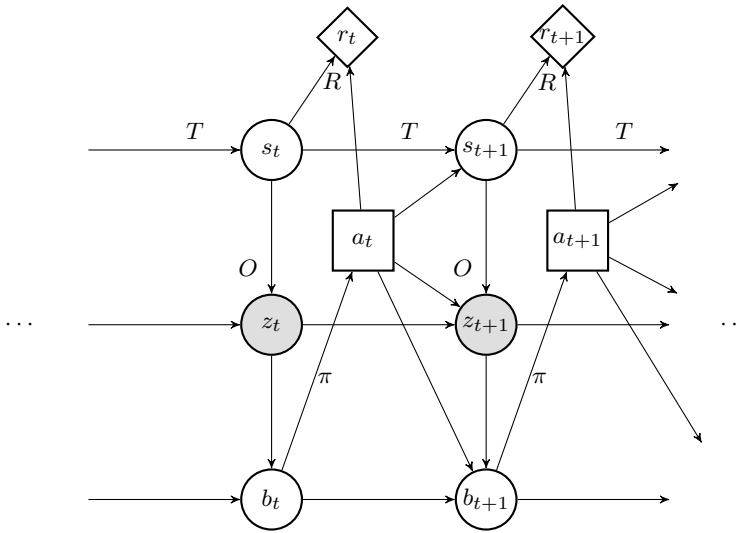
Finally, a POMDP can be represented as a special case of influence diagrams \citep{morato2020optimal, luque2019risk}, which form a class of probabilistic graphical models. Figure \ref{fig:pomdp} illustrates the influence diagram for the POMDP here considered. Circles and rectangles correspond to random and decision variables, respectively, while diamonds correspond to utility functions \citep{koller2009probabilistic}. Shaded shapes denote observed variables, while edges encode the dependence structure among variables.

\section{The railway maintenance problem}\label{sec:data}
We apply and test the proposed methodology on the problem of optimal maintenance planning for railway infrastructure assets on the basis of availability of regularly acquired monitoring data. The railway track comprises various components, such as rails, sleepers, and ballast, which are exposed to harsh environments and high operating loads, leading to accelerated degradation.
Among these infrastructure components, the substructure - in particular - is especially important in this degradation process. The substructure undergoes repeated loading from the superstructure (tracks, sleepers and ballast), prevents soil particles from rising into the ballast, and facilitates water drainage. A weakened substructure typically results in distortions of the track geometry.
Tamping \citep{Audley2013}, a maintenance procedure that uses machines to compact the ballast underneath the railway track, restoring its shape, stability and drainage system, is often applied when the substructure condition is considered moderately deteriorated. 
However, in case of poor substructure condition, such as intrusion of clay or mud or water clogging, tamping provides only a short-term remedy, and replacing the superstructure and substructure is the most appropriate long-term solution.

The optimization of maintenance decisions for these critical infrastructure components benefits from information that is additional to the practice of scheduled visual inspections, which are typically conducted on-site by experts. Such additional information can be delivered from monitoring data derived by diagnostic vehicles. In this work, we specifically exploit the \textit{fractal values}, a substructure condition indicator extracted from the longitudinal level, which is measured by a laser-based system mounted on a diagnostic vehicle, to guide decisions for substructure renewal.
The longitudinal level represents the deviations of the rail from a smoothed vertical position \citep{wang2021study}. 
On the basis of this measurement the fractal values can be computed, via appropriate filtering and processing steps. The fractal value indicator describes the degree of ``roughness" of the track at varying wavelength scales. For the interested reader, the detailed steps of the fractal value computation are reported in \citet{Landgraf2019, arcieri2022bridging}. In particular, long-wave (25-70~m) fractal values, which are employed in this work, have shown a significant correlation to substructure damage \citep{HoelzlIABMAS2021}, and are used by railway authorities as an indicator which can instigate repair/maintenance actions, such as tamping. 

In this work, we use actual track geometry measurements, carried out via a diagnostic vehicle of the SBB between 2008 and 2018 across Switzerland's railway network. The track geometry measurements were collected twice a year for the investigated portion of track. The fractal values are computed every 2.5m from the measured longitudinal level. The performed maintenance actions have been logged for the analysed tracks over the same considered period. These logs contain information on the maintenance, repair, or renewal actions taken on a section of the network at a specific date. 

We model the railway track maintenance optimization with a POMDP scheme, relying on diagnostic vehicle measurements of long-wave fractal values. The true but unobserved railway condition is discretized in 4 hidden states, $s_0$, $s_1$, $s_2$, and $s_3$, reflecting various grades, from perfect to highly deteriorated state. This is chosen to coincide with the number of grade levels assumed by the Swiss Federal Railways for classifying substructure condition. It should be noted, that in the POMDP inference setting, the number of hidden states is not fixed. To this end, we evaluated further possible dimensions of the hidden states vector, as part of the POMDP inference presented in the next section; a dimension of four yielded improved convergence and better-defined distributions. The fractal values are assumed as the (uncertain) POMDP observations, which correlate with the actual state of the substructure, but offer only partial and noisy information thereof. Unlike classical POMDP modeling of optimal maintenance planning problems, where observations are usually discrete, fractal values comprise (negative) continuous values, rendering the considered POMDP inference and solution quite complex. The problem definition is supplemented with information on the available maintenance actions. Three possible actions are considered, corresponding to the real-world setting, namely action $a_0$ do-nothing, and the aforementioned tamping and replacement actions, denoted as $a_1$ and $a_2$, which can be interpreted as a minor and a major repair, respectively. The fractal value indicators are derived via measurements of the diagnostic vehicle every 6 months, which thus represents the time-step of the decision-making problem. Considering the almost 10 years of collected measurements, our real-world dataset is ultimately composed of time-series of 20 fractal values, per considered railway section, complete with information on respective maintenance actions (with ``action" do-nothing included), i.e., $(z_0, a_0, \cdots, a_{19}, z_{20})$. Finally, the (negative) rewards representing costs associated with actions and states have been elicited from SBB and are reported in Table \ref{tab:costs} in general cost units.

\begin{table}[!h]
\centering
\caption{Costs of the POMDP model.}
    \centering
    \begin{tabular}{ccccc}
        \multicolumn{1}{c}{\bf State condition}  &\multicolumn{1}{c}{\bf $s_0$}  &\multicolumn{1}{c}{\bf $s_1$} &\multicolumn{1}{c}{\bf $s_2$}
        &\multicolumn{1}{c}{\bf $s_3$}\\
        \hline
        \bf Maintenance action\\
        $a_0$ & $0$   & $0$   & $0$ & $0$ \\
        $a_1$ & $-50$ & $-50$ & $-50$ & $-50$ \\
        $a_2$ & $-2,050$ & $-2,710$ & $-3,370$ & $-4,050$ \\
        \textbf{Condition cost} & $-100$ & $-200$ & $-1,000$ & $-8,000$\\
    \end{tabular}
    \label{tab:costs}
\end{table}

\section{POMDP inference}\label{sec:inference}
To formulate the POMDP problem, the transition dynamics and the observation generating process must be inferred. In the RL context, the POMDP inference is necessary to generate samples for the policy learning, for inference of a belief over the hidden states, and/or for testing purposes. To tackle this key issue, we propose an MCMC inference of a HMM conditioned on actions, which jointly estimates parameter distributions of both the POMDP transition and observation models based on available data. While we implement the proposed scheme on the problem of railway maintenance planning based on fractal value observations, its applicability is general. Therefore, we further suggest possible extensions to help researchers and practitioners tailor the POMDP model inference to the problem at hand. In addition, we provide a complementary tutorial\footnote{Code available on \href{https://github.com/giarcieri/Hidden-Markov-Models}{GitHub}.} illustrating  the code implementation on various simulated case-studies, in order to support exploitation for real-world applications.

In the context of discrete hidden states and actions, the transition dynamics are modelled via Dirichlet distributions:
\begin{align}\label{eq:transition-model}
\begin{split}
    T_0 & \sim \dirichlet(\alpha_0)\\
    s_0 & \sim \categorical(T_0)\\
    T & \sim \dirichlet(\alpha_T)\\
    s_t \mid s_{t-1}, a_{t-1} & \sim \categorical(T)\\
\end{split}
\end{align}
where $T_0$ are the parameters of the probability distribution of the initial state $s_0$, and $\alpha_0$ and $\alpha_T$ are the prior concentration parameters. $T_0$ can be assigned a uniform flat prior $\alpha_0$, unless some prior knowledge on the initial state distribution is available. By contrast, it is beneficial to regularize $T$ with informative priors $\alpha_T$, which regularize the deterioration or the repairing process. For example, the transition matrix related to the action do-nothing, which describes the deterioration process of the system, can be regularized with higher prior probabilities on the diagonal and on the upper-right triangle, and near-zero on the lower-left triangle. Likewise, the transition matrices associated with maintenance actions would present higher prior probabilities on the left triangle and near-zero on the right triangle, in order to inform the model that a repair action is expected to be followed by improvements of the system. 

The dimensionality of the Dirichlet distribution that models the transition dynamics $T$ is $S \times S \times A$, namely one transition matrix per action. The extension to time-dependent transition dynamics is straightforward by enlarging the distribution by a further dimension representing time, i.e., $S \times S \times A \times H$.

In the context of continuous observations, the observation generating process can differ on the basis of whether the observation follows a deterioration or a repairing process. In addition, similarly to the inference of the first hidden state according to $T_0$, an initial observation process can be necessary to model the first observation. Tailoring to the nature of the fractal value monitoring data, the initial, deterioration, and repairing processes are modelled via Truncated Student's $t$ processes, as follows:
\begin{align}\label{eq:obs-model}
\begin{split}
    z_0  & \sim \TS(\mu_{s_{t_0}}, \sigma_{s_{t_0}}, \nu_{s_{t_0}}, \ub=0)\\
    z_t - z_{t-1} & \sim \TS(\mu_{d\mid s_t}, \sigma_{d\mid s_t}, \nu_{d\mid s_t}, \ub=- z_{t-1})\\
    z_t  & \sim \TS(k_{r\mid a_{t-1}}*z_{t-1}+\mu_{r\mid s_t}, \sigma_{r\mid s_t}, \nu_{r\mid s_t}, \ub=0)\\
\end{split}
\end{align}
where $\ub$ stands for ``upper bound'', and all parameters governing the processes are assigned priors described in \citet{arcieri2022bridging}. 

The use of Truncated Student's $t$ processes was tailored to the mathematical characteristics of the fractal values, which i) assume only negative values, ii) exhibit a negative trend in absence of repairing actions, iii) their values are dependent on the previous observations, and iv) the studied dataset, as is common in real-world measurements, presents outliers and measurement errors, modelled by the Student's $t$ fat tails. Naturally, other distributions can also be employed as part of the proposed framework in order to model the data at hand related to each application. For instance, in absence of the previous limiting characteristics, simpler (unbounded) Gaussian emissions could have been used, as further shown in the tutorial. In the case of discrete observations, the observation model would be represented by a probability matrix $S \times Z$, which can be again modelled via a Dirichlet distribution. In the case of more than one possible inspection action or monitoring tool, as in \citet{papakonstantinou2018pomdp}, the Dirichlet distribution can be simply enlarged by a further dimension representing the number of possibilities. Finally, dependencies in multi-component systems could be modelled via a Bayesian hierarchical model \citep{gelman1995bayesian}, enabling solutions as proposed in \cite{andriotis2019managing, andriotis2021deep, morato2023inference}.

\begin{figure}[!ht]
\centering
\resizebox{7.5cm}{9cm}{
\begin{tikzpicture}[every text node part/.style={align=center}]
    \node [latent] (Z) at (0,0) {Observation model\\$\sim$\\$\TS$};
    \node [latent,above=of Z] (s) {Hidden states\\$\sim$\\$\categorical$};
    \node [latent,above=of s] (T){Transition model\\$\sim$\\$\dirichlet$};
    \node [obs, right=of s] (a) {Actions};
    \node [rectangle,obs, left=of Z] (Obs) {Observations\\(Fractal values)};
    
    \path [draw,->] (T) edge (s);
    \path [draw,->] (s) edge (Z);
    \path [draw,->] (Obs) edge (Z);
    \path [draw,->] (a) edge (Z);
    \path [draw,->] (a) edge (s);

    \plate [inner sep=.25cm,xshift=.12cm,yshift=.2cm] {plate1} {(Z)(s)(T)} {Model inference};

\end{tikzpicture}
}
\caption{A graphical model of the HMM inference. Arrows indicate dependencies, while shaded nodes indicate observed variables.}
\label{fig:hmm}
\end{figure}
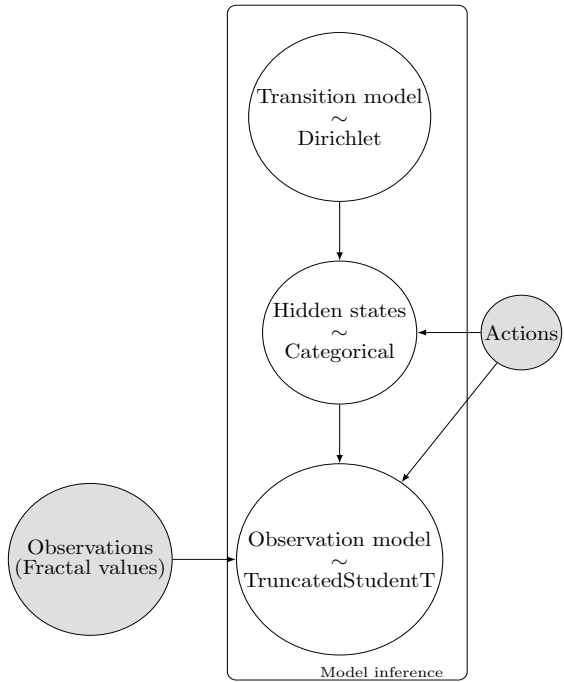

The graphical model of the entire HMM is reported in Figure \ref{fig:hmm}. The MCMC inference is run on a final dataset of 62 time-series with the No-U-Turn Sampler (NUTS) \citep{hoffman2014no}. Four chains are run with 3,000 samples collected per chain. The inference results, which present good post-inference diagnostic statistics, with no divergences and high homogeneity between and within chains, are reported in Figures \ref{fig:tr_mat0}-\ref{fig:params_i} in Appendix \ref{secA1}.

\section{RL for POMDP solution}\label{sec:ComparisonSolution}

POMDP problems have been tackled via deep RL with common methods augmented with LSTM architectures and a history of past observations (and possibly actions) as inputs \citep{zhu2017improving, meng2021memory}. More recently, motivated by the breakthrough success of Transformers over LSTMs in natural language processing, \citet{parisotto2020stabilizing} designed a new transformer architecture, namely GTrXL, which yielded significant improvements in terms of performance and robustness over LSTMs on a set of partially observable benchmarking tasks. A main advantage of GTrXL is the capability to vary the dimensionality of the input over time. While LSTMs generally require a fixed window of $h$ past observations, requiring the use of dummy observations in the first $h-1$ decision time-steps, the GTrXL can at every time-step base the decisions on the entire history of past observations (and actions).

Both LSTM and GTrXL architectures compose fully model-free deep RL solutions to POMDPs. A third modeling option, which comprises a model-based/model-free hybrid solution, pertains to transformation of the POMDP problem into the belief-MDP by computing beliefs via Bayes Theorem (Equation \ref{eq:belief}). The belief-MDP is then solved via classical deep model-free RL methods with feed-forward NNs \citep{andriotis2019managing, morato2023inference}. We here compare the performance of the two model-free and the hybrid solution, referred to as ``belief-input'' case, on the real-world POMDP problem of railway maintenance planning that has been presented in Section \ref{sec:data}, with parameter inference described in Section \ref{sec:inference}. While \citet{parisotto2020stabilizing} demonstrate the superiority of Transformers over LSTMs on simulated tasks, our work offers a further comparison of the two methods, and confirms the superiority of the former, on a real-world stochastic (both in the transition dynamics and in the observation generating process), partially observable problem.

For this comparison we set the POMDP parameters to the mean values of the distributions reported in Appendix \ref{secA1}, in order to evaluate the methods without model uncertainty, with the latter case tackled in the next section. For all modeling options, the policy is learned via the Proximal Policy Optimization (PPO) algorithm with clipped surrogate objective \citep{schulman2017proximal}. The overall evaluation algorithm is reported in pseudocode format in Algorithm \ref{alg:evaluation}. In addition, the code of the experiment is made available online\footnote{Code available on \href{https://github.com/giarcieri/Robust-optimal-maintenance-planning-through-reinforcement-learning-and-rllib}{GitHub}.}. We consider 50 time-steps, i.e., 25 years (1 time-step equals 6 months), as the decision horizon $H$ of the problem, as discussed with our SBB partners.

\begin{algorithm}[!ht]
\caption{Evaluation algorithm}\label{alg:evaluation}
\begin{algorithmic}[1]
\State{Initialize policy network $\pi_{\phi}$}
\State{Initialize replay buffer $\mathcal{D} \leftarrow \emptyset$}
\State{Set environment parameters $\hat{\theta}$ to the mean values of $p(\theta\mid D)$}
\For{training episode = 0 to $N$}
\State{Sample initial $s_0 \sim T_{0_{\hat{\theta}}}$ and $z_0 \sim O_{0_{\hat{\theta}}}$}
\State{Initialize belief to initial state distribution $b_0\gets T_{0_{\hat{\theta}}}$}
    \For{timestep t = 0 to $H$}
        \If{belief-input case}
            \State{Input $y_t = b_t$}
        \ElsIf{LSTM}
            \State{Input $y_t = (z_t, a_{t-1},\cdots, z_{t-h+1})$} \Comment{$h=3$}
        \ElsIf{GTrXL}
            \State{Input $y_t = (z_t, a_{t-1},\cdots, z_0)$}
        \EndIf
        \State{$a_t \sim \pi_{\phi}(y_t)$}
        \State{$s_{t+1} \sim T_{\hat{\theta}}(s_t, a_t), z_{t+1} \sim O_{\hat{\theta}}(s_{t+1}, a_t, z_{t})$}
        \State{Compute $b_{t+1}$ via Equation \ref{eq:belief}}
        \State{$\mathcal{D} \leftarrow \mathcal{D} \cup\left\{\left(y_{t}, a_{t}, R\left(s_{t}, a_{t}\right)\right)\right\}$}
    \EndFor
    \State{\textbf{every} $K$ total timesteps \textbf{do}} \Comment{$K=4,000$}
    \State{\hspace*{\algorithmicindent}Update $\pi_{\phi}$ with PPO and replay buffer $\mathcal{D}$}
    \State{\textbf{every} $5$ updates \textbf{do}}
    \State{\hspace*{\algorithmicindent}Run 500 policy evaluation episodes without exploration}
\EndFor
\end{algorithmic}
\end{algorithm}

\begin{figure}[!ht]
    \centering
    \begin{tikzpicture}
    \node (img)  {\includegraphics[width=1\textwidth]{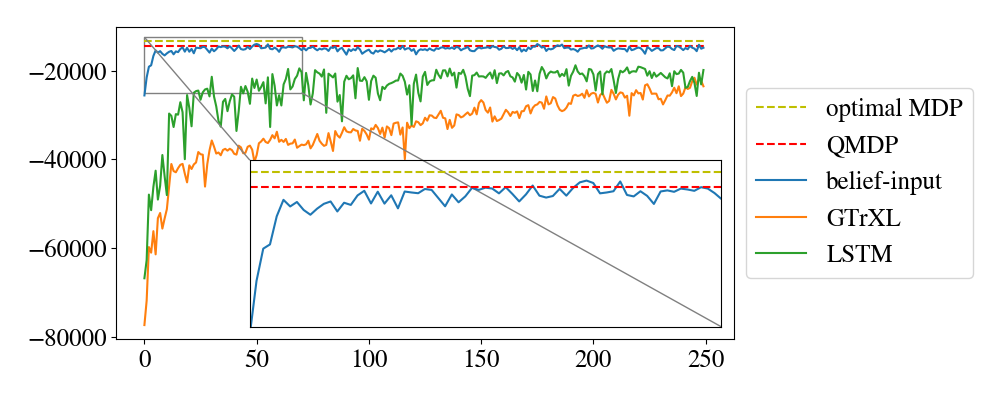}};
    \node[below=of img, node distance=0cm,xshift=-0.4cm,  yshift=1.5cm] {Evaluation iteration};
    \node[left=of img, node distance=0cm, rotate=90, anchor=center,xshift=0.2cm,yshift=-1.2cm] {Total costs};
    \end{tikzpicture}
    \caption{Comparison of the performance of LSTM (green), GTrXL (orange), and the belief-input case (blue) over 250 evaluation iterations. At every iteration, 500 trial episodes are evaluated with different random seeds and the average results are returned. The entire analysis is repeated for a second random seed and the average performance is plotted. An evaluation iteration is run after 5 policy updates and a policy update is performed every 4,000 training time-steps, for a total of 5 million time-steps. The performance is further benchmarked against the $Q_{MDP}$ method (dashed red) and the optimal MDP policy (dashed yellow). On the left corner, a zoomed-in plot of the belief-input performance over the first 70 evaluation iterations.}
    \label{fig:comparison_250}
\end{figure}

For all methods, the policy networks are updated every 4,000 training time-steps. Every 5 updates, 500 evaluation episodes are run with different random seeds in order to average the results over the stochasticity of the environment. In addition, the entire analysis is repeated a second time (with a different random seed) to further average the results over the stochasticity of the NN training. Grid-searches are performed over the hyperparameters for all methods and the selected values are reported in Table \ref{tab:hyper} in Appendix \ref{app:hyper}. The average performance over 250 evaluation iterations (5 million training time-steps) is plotted in Figure \ref{fig:comparison_250}. Along with the three evaluated methods, two additional benchmarking solutions are reported. The first option refers to the $Q_{MDP}$ method \citep{littman1995learning}, which constitutes a POMDP solution based on DP, and which turns out to be an effective solution for the characteristics of this problem \citep{arcieri2022bridging}. The second option is the optimal MDP solution, namely the optimal policy computed and evaluated on the underlying MDP, i.e., when the hidden states are fully observable. The latter constitutes an upper bound to any POMDP solution, which cannot be exceeded, given the irreducible inherent uncertainty of the observations, and serves as a benchmarking reference.

\begin{figure}[!ht]
    \centering
    \begin{tikzpicture}
    \node (img)  {\includegraphics[width=1\textwidth]{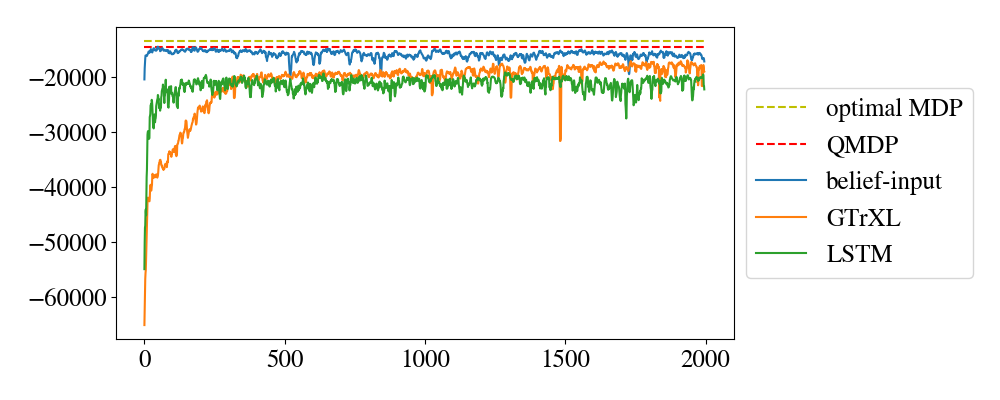}};
    \node[below=of img, node distance=0cm,xshift=-0.4cm,  yshift=1.5cm] {Evaluation iteration};
    \node[left=of img, node distance=0cm, rotate=90, anchor=center,xshift=0.2cm,yshift=-1.2cm] {Total costs};
    \end{tikzpicture}
    \caption{Comparison of the performance of LSTM (green), GTrXL (orange), and the belief-input case (blue) over 2,000 evaluation iterations, for a total of 40 million training time-steps. The performance is further plotted with an average rolling window of 5 steps for displaying purposes.}
    \label{fig:comparison}
\end{figure}

The belief-input method outperforms the other two model-free RL solutions and already shows strong performance at the first evaluation iterations. The method converges to the best policy within a few iterations, as reported in the zoomed-in view of the first 70 evaluation iterations reported in the lower left figure inset, matching the $Q_{MDP}$ method with few policy updates. Because the number of training time-steps evaluated may not be sufficient for convergence of the other two model-free RL methods, we continue training up to 2,000 evaluation iterations (40 million training time-steps). This could however negatively impact the performance of the belief-input method, which already converged and may begin to suffer from overfitting. The extended training is reported in Figure \ref{fig:comparison}, where a rolling average window of 5 steps is further applied for illustration purposes.

As expected, the performance of the belief-input method slightly decreases over time. The GTrXL is proven to deliver a better architecture than the LSTM for POMDP applications, also for this particular case of application on a real-world problem. The GTrXL is indeed less affected by variance and eventually converges to a better policy, albeit still far from the $Q_{MDP}$ benchmark and the best policy with the belief-input method.

Finally, for all three methods we saved the best models, which were evaluated during training and evaluated the learned policies over 100,000 trials. The results are reported in Table \ref{tab:comparison} in terms of average performance, Standard Error (SE), best (Max) and worst (Min) trial. In the table, the belief-input case average performance is close but slightly worse than the $Q_{MDP}$ method. This is likely due to the fact that the best model was picked based on an average over 500 trials, which is still subject to a significant standard error.

\begin{table}[h]
\begin{center}
\caption{Performance of the best models inferred during the training process, evaluated over 100,000 simulations.}\label{tab:comparison}
\begin{tabular}{@{}lcccc@{}}
\toprule
Method & Avg. performance  & SE & Max & Min\\
\midrule
Optimal MDP  & -13,315   & 27  & -5,000 & -93,980 \\
$Q_{MDP}$       & -14,374   & 35  & -5,050 & -123,800 \\
Belief-input & -14,677   & 36  & -5,050  & -121,950 \\
GTrXL        & -17,196   & 46  & -5,700  & -188,600 \\
LSTM         & -18,167  & 42 & -5,100  & -404,150 \\
\botrule
\end{tabular}
\end{center}
\end{table}

\section{Domain randomization for robust solution}\label{sec:DomainRandomization}
Further to the challenge of POMDP inference, another key issue is the robustness of the deep RL solutions. RL methods generally learn an optimal policy by interacting with a simulator. When the trained RL agent is deployed to the real-world, the performance can deteriorate, or altogether fail, due to the ``simulation-to-reality'' gap \citep{zhao2020sim, salvato2021crossing}, if the solution is not robust to model uncertainty. 

In \citet{arcieri2022bridging}, we propose a framework in combination with the POMDP inference to enhance the robustness of DP solutions to model uncertainty. Namely, the POMDP parameter distributions inferred via MCMC sampling are incorporated into the solution by merging DP algorithms with Bayesian decision making. In Bayesian decision theory \citep{berger2013statistical}, given a utility function $U(\mathbf{\theta}, a)$ that maps possible outcomes to their utility, the parameters $\mathbf{\theta}$ of the problem, and some decision $a$, the Bayesian optimal action is the one which maximizes the expected utility with respect to parameter uncertainty:
\begin{equation}\label{eq:bdm}
    a^*=\argmax_{a\in A}\E_{\mathbf{\theta}\sim p(\mathbf{\theta})}\left[U(\mathbf{\theta}, a)\right]
\end{equation}
In \citet{arcieri2022bridging} we incorporate DP methods into Equation \ref{eq:bdm} to derive solutions that maximize the expected value with respect to the entire model parameter distributions, hence rendering the solution robust to model uncertainty.

\begin{figure}[!ht]
  \centering
  \includegraphics[width=\linewidth]{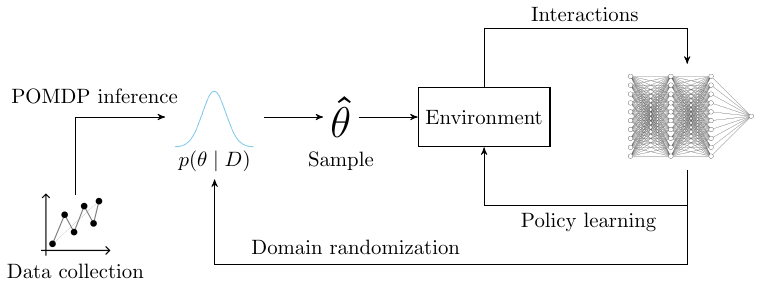}
  \caption{The POMDP inference and robust solution framework via domain randomization and deep reinforcement learning.}
  \label{fig:framework}
\end{figure}

In this work, we bring this framework into the RL training scheme. The utility function is represented by the RL algorithm objective function, e.g., the PPO clipped surrogate objective in this case. We propose the use of domain randomization \citep{tobin2017domain} of the POMDP environment, which is enabled by our POMDP inference scheme through the recovery of parameter distributions, in order to enhance the robustness of the RL solution to model uncertainty. At every episode, a different POMDP configuration is sampled from the parameter distributions. The RL agent interacts with this POMDP configuration until the end of the episode. Afterwards, a new configuration of the environment is sampled. At the end of the training, the RL agent will have optimized the learned policy over all possible problem parameters to derive a solution robust to model uncertainty. The expectation in Equation \ref{eq:bdm} is thus implemented in practice via stochastic gradient ascent/descent steps over varying randomized  problem parameters. It should be reminded that the (Bayesian) robust optimal policy may be sub-optimal for a specific value $\mathbf{\theta}$, while maximizing the expected value with respect to the entire model parameter distribution. The domain randomization technique can thus be used in combination with the model inference proposed in Section \ref{sec:inference} to establish a joint framework of POMDP inference and robust solution based on RL. The framework is depicted in the graphical model in Figure \ref{fig:framework}.

\begin{figure}[!ht]
    \centering
    \begin{tikzpicture}
    \node (img)  {\includegraphics[width=1\textwidth]{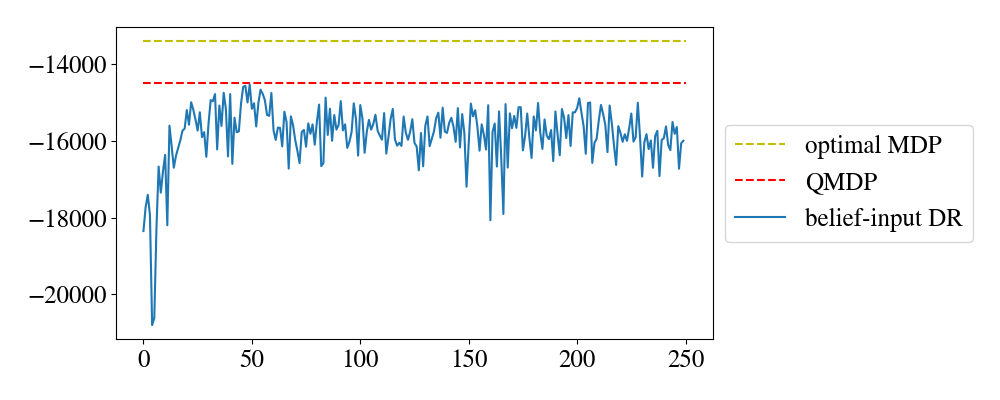}};
    \node[below=of img, node distance=0cm,xshift=-0.6cm,  yshift=1.5cm] {Evaluation iteration};
    \node[left=of img, node distance=0cm, rotate=90, anchor=center,xshift=0.2cm,yshift=-1.2cm] {Total costs};
    \end{tikzpicture}
    \caption{Performance of the belief-input case (blue) over 250 evaluation iterations with domain randomization, i.e., a different POMDP model is sampled at every episode, both for training and evaluation. At every iteration, 500 trial episodes are evaluated with different random seeds and the average results are returned. The entire analysis is repeated for a second random seed and the average performance is plotted. An evaluation iteration is run after 5 policy updates and a policy update is performed every 4,000 training time-steps, for a total of 5 million time-steps. The performance is further benchmarked against the robust $Q_{MDP}$ method (dashed red) and the robust optimal MDP policy (dashed yellow), evaluated under model uncertainty as in \citet{arcieri2022bridging}.}
    \label{fig:belief-dr}
\end{figure}

We showcase the implementation of this framework with the belief-input method, but it is also applicable with the other methods reported in Table \ref{tab:comparison} given its general validity. The evaluation algorithm is similar to Algorithm \ref{alg:evaluation}, with the only difference that the POMDP parameters $\hat{\theta}$ are sampled at every episode from the inferred posterior distributions $p(\theta\mid D)$. The policy updates are again performed every 4,000 training time-steps and an evaluation iteration is run every 5 policy updates. Similarly to Figure \ref{fig:comparison}, the performance during training is averaged at each evaluation iteration over 500 episodes with different random seeds. The analysis is then repeated for a second random seed to also average over the stochasticity of the NN training. The resulting average performance is plotted in Figure \ref{fig:belief-dr}. Given the more challenging learning task, owing to model uncertainty, the average training performance decreases and demonstrates a higher variance than the belief-input performance without domain randomization, shown in Figure \ref{fig:comparison}. For this case, the hyper-parameter tuning was also restricted to a minimal grid-search. While the results are already satisfying, the RL agent performance can likely be further increased via a more thorough hyperparameter optimization.

Again, the best performing models shown in the evaluations during training are saved and the learned policy is evaluated over 100,000 simulations. The results are shown in Table \ref{tab:DR} and compared against the robust $Q_{MDP}$ policy described in \citet{arcieri2022bridging} and the upper bound optimal MDP policy evaluated with full observability, both assessed under model uncertainty. In addition, we report the result of the best model of the RL agent from the previous analysis, namely with the policy optimized without model uncertainty incorporated into the training (i.e., no domain randomization), evaluated now in the context of model uncertainty. This further analysis resembles a real-world deployment, where the environment parameters can differ from those inferred, inducing the aforementioned simulation-to-reality gap. The performance of the agent trained with no domain randomization deteriorates, while the agent trained with domain randomization is able to learn and deliver a more robust policy in the context of model uncertainty.

\begin{table}[h]
\begin{center}
\caption{Performance of the best models during training evaluated over 100,000 simulations in the context of model uncertainty with domain randomization. In particular, we report on the evaluation of the belief-input agent trained with (DR) and without Domain Randomization (no DR). The former achieves a significantly improved and more robust policy.}\label{tab:DR}
\begin{tabular}{@{}lcccc@{}}
\toprule
Method & Avg. performance  & SE & Max & Min\\
\midrule
Optimal MDP  & -13,374   & 33  & -5,000 & -190,450 \\
$Q_{MDP}$       & -14,526   & 39  & -5,050 & -197,050 \\
Belief-input DR & -14,648   & 38  & -5,050  & -168,600 \\
Belief-input no DR & -14,901   & 39  & -5,050  & -205,100 \\
\botrule
\end{tabular}
\end{center}
\end{table}

\begin{figure}
     \centering
     \begin{subfigure}[b]{0.49\textwidth}
         \centering
         \begin{tikzpicture}
         \node (img)  {\includegraphics[width=\textwidth]{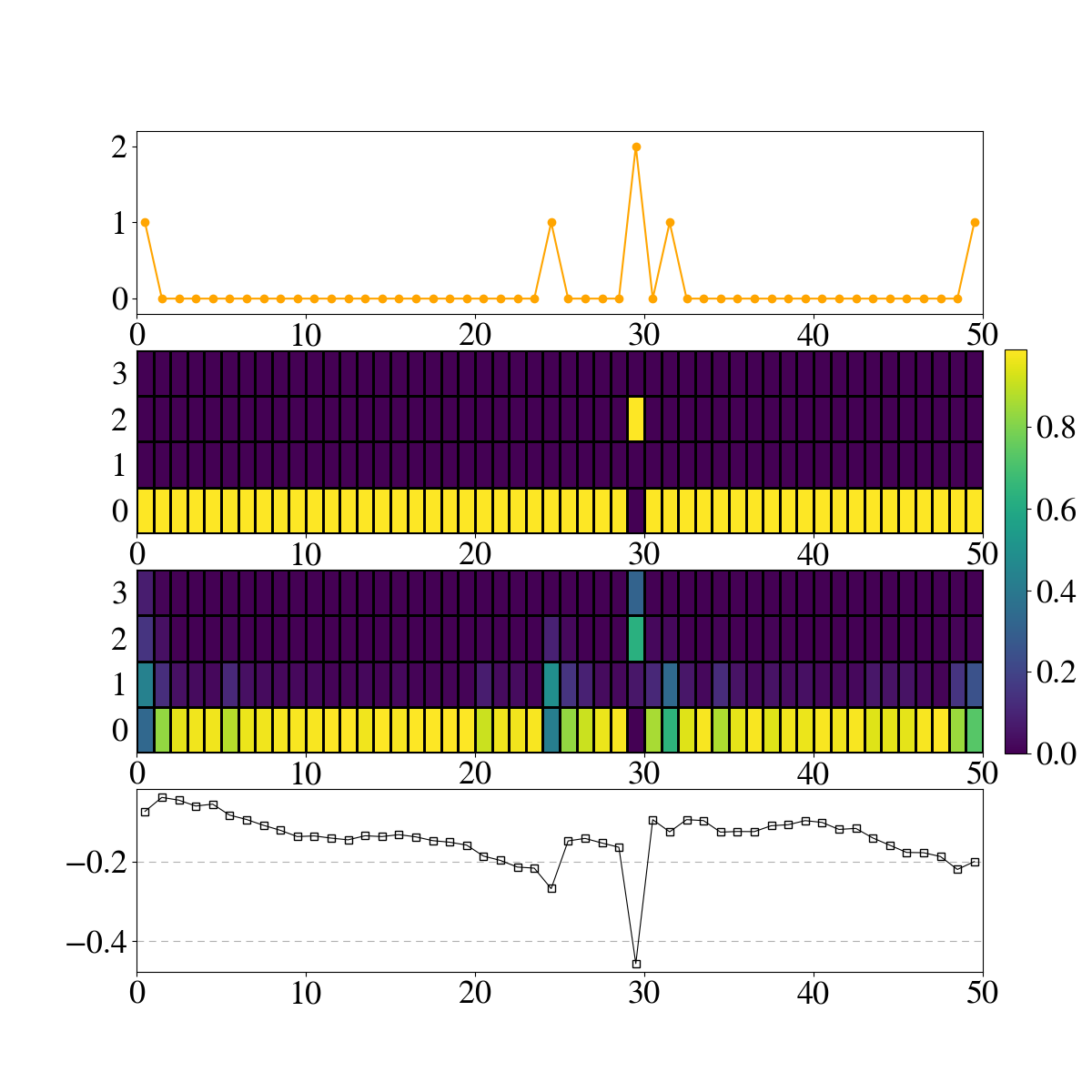}};
          \node[below=of img, node distance=0cm,xshift=0cm,  yshift=1.6cm] {\footnotesize Timestep};
          \node[left=of img, node distance=0cm, rotate=90, anchor=center,xshift=1.7cm, yshift=-1.35cm] {\footnotesize Action};
          \node[left=of img, node distance=0cm, rotate=90, anchor=center,xshift=0.5cm, yshift=-1.35cm] {\footnotesize State};
          \node[left=of img, node distance=0cm, rotate=90, anchor=center,xshift=-0.6cm,yshift=-1.35cm] {\footnotesize Belief};
          \node[left=of img, node distance=0cm, rotate=90, anchor=center,xshift=-1.85cm,yshift=-1.35cm] {\footnotesize Obs};
          \end{tikzpicture}
     \end{subfigure}
     \begin{subfigure}[b]{0.49\textwidth}
         \centering
         \begin{tikzpicture}
         \node (img)  {\includegraphics[width=\textwidth]{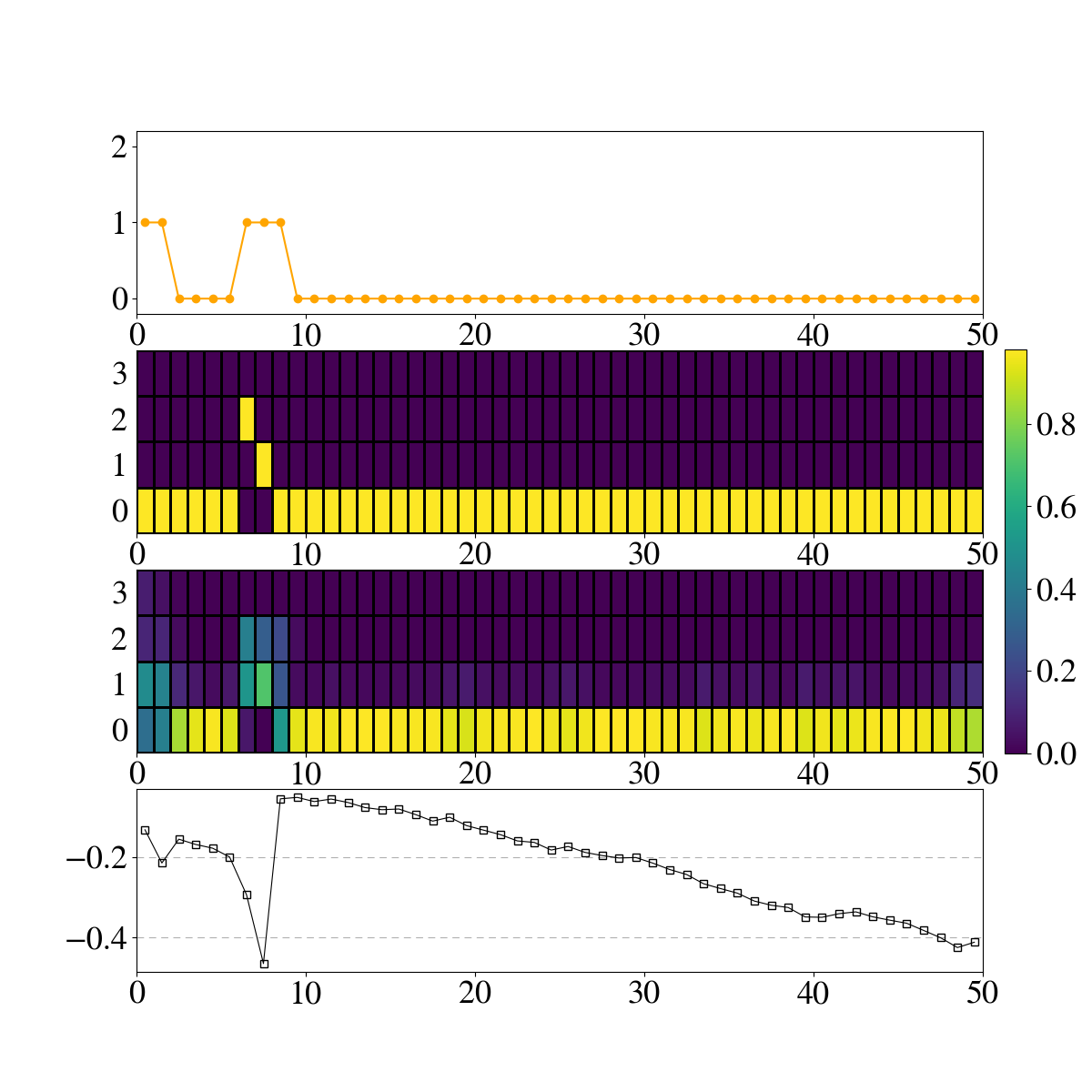}};
         \node[below=of img, node distance=0cm,xshift=0cm,  yshift=1.6cm] {\footnotesize Timestep};
          \node[left=of img, node distance=0cm, rotate=90, anchor=center,xshift=1.7cm, yshift=-1.35cm] {\footnotesize Action};
          \node[left=of img, node distance=0cm, rotate=90, anchor=center,xshift=0.5cm, yshift=-1.35cm] {\footnotesize State};
          \node[left=of img, node distance=0cm, rotate=90, anchor=center,xshift=-0.6cm,yshift=-1.35cm] {\footnotesize Belief};
          \node[left=of img, node distance=0cm, rotate=90, anchor=center,xshift=-1.85cm,yshift=-1.35cm] {\footnotesize Obs};
          \end{tikzpicture}
     \end{subfigure}
        \caption{Two trials of the maintenance actions planned by the belief-input model trained with domain randomization. From bottom to top: the observations (fractal values); the beliefs, namely a probability distribution over hidden states, computed via Bayes' formula and fed to the policy networks; the true hidden states, which are not accessed by the agent and/or the model; the actions planned by the RL agent.}
        \label{fig:trials}
\end{figure}
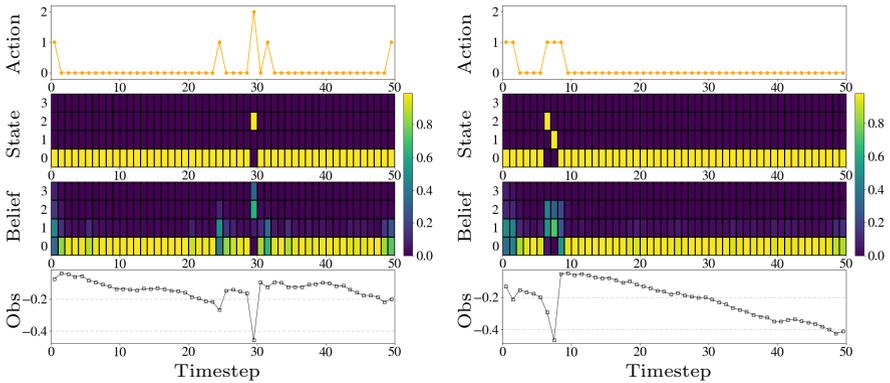

Finally, Figure \ref{fig:trials} shows two trials of the maintenance actions planned by the belief-input model, which has been trained with domain randomization. From bottom to top: the observations (fractal values); the beliefs, namely the probability distribution over hidden states, computed via Bayes' formula and fed to the policy networks; the true hidden states, which are not accessed by the agent and/or the belief computations; the actions planned by the RL agent.

\section{Conclusion}\label{sec:conc}
This work tackles two key issues relating to adoption of RL applications in real-world partially observable planning problems. Firstly, a POMDP model, which enables the RL training via simulations, is often unknown and generally non-trivial to infer, with unified best practices not available in the literature. This constitutes a main obstacle against broad adoption of the POMDP scheme and its solution methods for real-world applications. Second, RL solutions often lack robustness to model uncertainty and suffer from the simulation-to-reality gap.

In this work, we tackle both issues via a combined framework for inference and robust solution of POMDPs based on deep RL algorithms. The POMDP inference is carried out via MCMC sampling of a HMM conditioned on actions, which jointly estimates the full distributions of plausible values of the transition and observation model parameters. Then, the parameter distributions are incorporated into the solution via domain randomization of the environment, enabling the RL agent to learn a policy, which is optimized over the space of plausible problem parameters and is, thus, robust to model uncertainty. We compare three common RL modeling options, namely a Transformer and an LSTM-based approach, which constitute model-free RL solutions, and a hybrid belief-input case. We implement our methods for optimal maintenance planning of railway tracks based on real-world monitoring data. While the Transformer delivers generally better performance than the LSTM, both methods are significantly outperformed by the hybrid belief-input case. In addition, we demonstrate on the latter method that an RL agent trained with domain randomization is able to learn an improved policy, which is robust to model uncertainty, than an RL agent trained without domain randomization.

A possible limitation of this work is that, while our methods allow for incorporation of rather complex extensions, e.g., time-dependent dynamics and hierarchical components, and are here demonstrated on the quite difficult case of continuous observations, the POMDP inference under continuous multi-dimensional states and actions is still to be investigated. Future work will focus on the development of methods that can scale to these cases, e.g, via coupling with deep model-based RL methods \citep{arcieri2021model}.

\backmatter

\bmhead{Acknowledgments}
The authors acknowledge the support of the Swiss Federal Railways (SBB) as part of the ETH Mobility Initiative project REASSESS. The authors thank the ETH cluster support for their precious help with the availability of computational power.

\section*{Declarations}

\subsection*{Funding}
The authors acknowledge the support of the Swiss Federal Railways (SBB) as part of the ETH Mobility Initiative project REASSESS.

\subsection*{Conflicts of interest/Competing interests}
The authors have no competing interests to declare that are relevant to the content of this article.

\subsection*{Ethics approval}
Not applicable.

\subsection*{Consent to participate}
Not applicable.

\subsection*{Consent for publication}
Not applicable. No further consent is needed for publication of this research paper.

\subsection*{Availability of data and material}
The real-world monitoring data used in this research paper is SBB proprietary and cannot be published.

\subsection*{Code availability}
All code of the experiments of this research paper is made available on GitHub in public repositories linked in the paper.

\subsection*{Authors' contributions}
\begin{itemize}
    \item Giacomo Arcieri: Conceptualization; Data curation; Formal analysis; Investigation; Methodology; Software; Visualization; Roles/Writing - original draft; Writing - review \& editing.
    \item Cyprien Hoelzl: Data curation; Roles/Writing - original draft.
    \item Oliver Schwery: Funding acquisition; Validation.
    \item Daniel Straub: Methodology; Supervision; Validation; Writing - review \& editing.
    \item Konstantinos G. Papakonstantinou: Methodology; Supervision; Validation; Writing - review \& editing.
    \item Eleni Chatzi: Conceptualization; Methodology; Funding acquisition; Project administration; Resources; Supervision; Validation; Writing - review \& editing.

\end{itemize}

\newpage

\begin{appendices}

\section{Inference results}\label{secA1}

\subsection{Transition model parameters}
\label{app:transition_observation}

\begin{figure}[!ht]
\centering
    \centering
    \includegraphics[width=\linewidth]{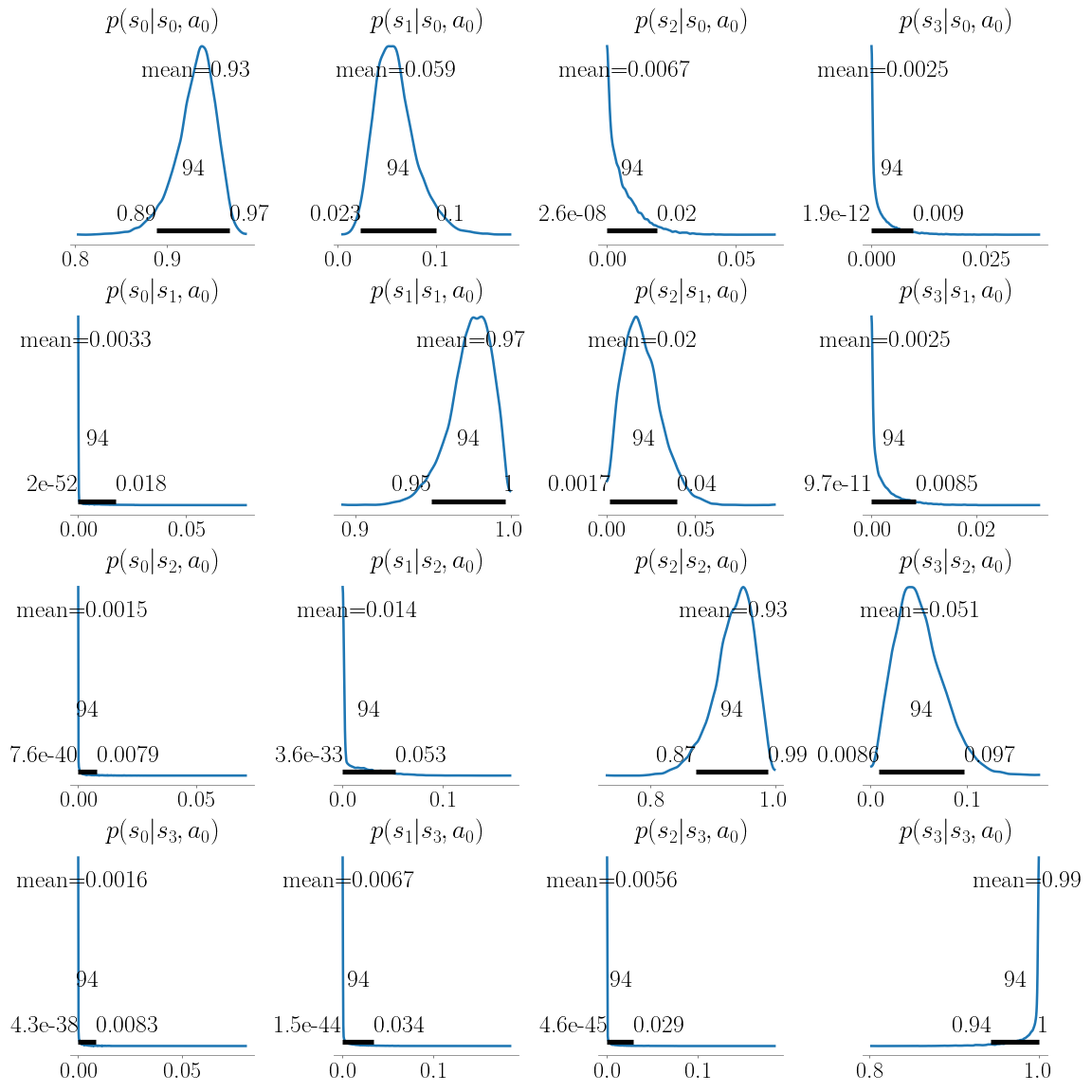}
    \caption{Transition matrix related to action do-nothing $a_0$. The distribution at row $i$ and column $j$ is associated with the probability to transition from state $i$ to $j$ when action $a_0$ is taken. Consistent with what is expected in deterioration processes the highest probabilities are assigned to the state remaining invariant (diagonal entries), lower probabilities exist for deterioration transitions (upper right triangle), and almost zero probability is assigned to improvements of the system (lower left triangle).}
    \label{fig:tr_mat0}
\end{figure}

\newpage

\begin{figure}[!ht]
    \centering
    \includegraphics[width=\linewidth]{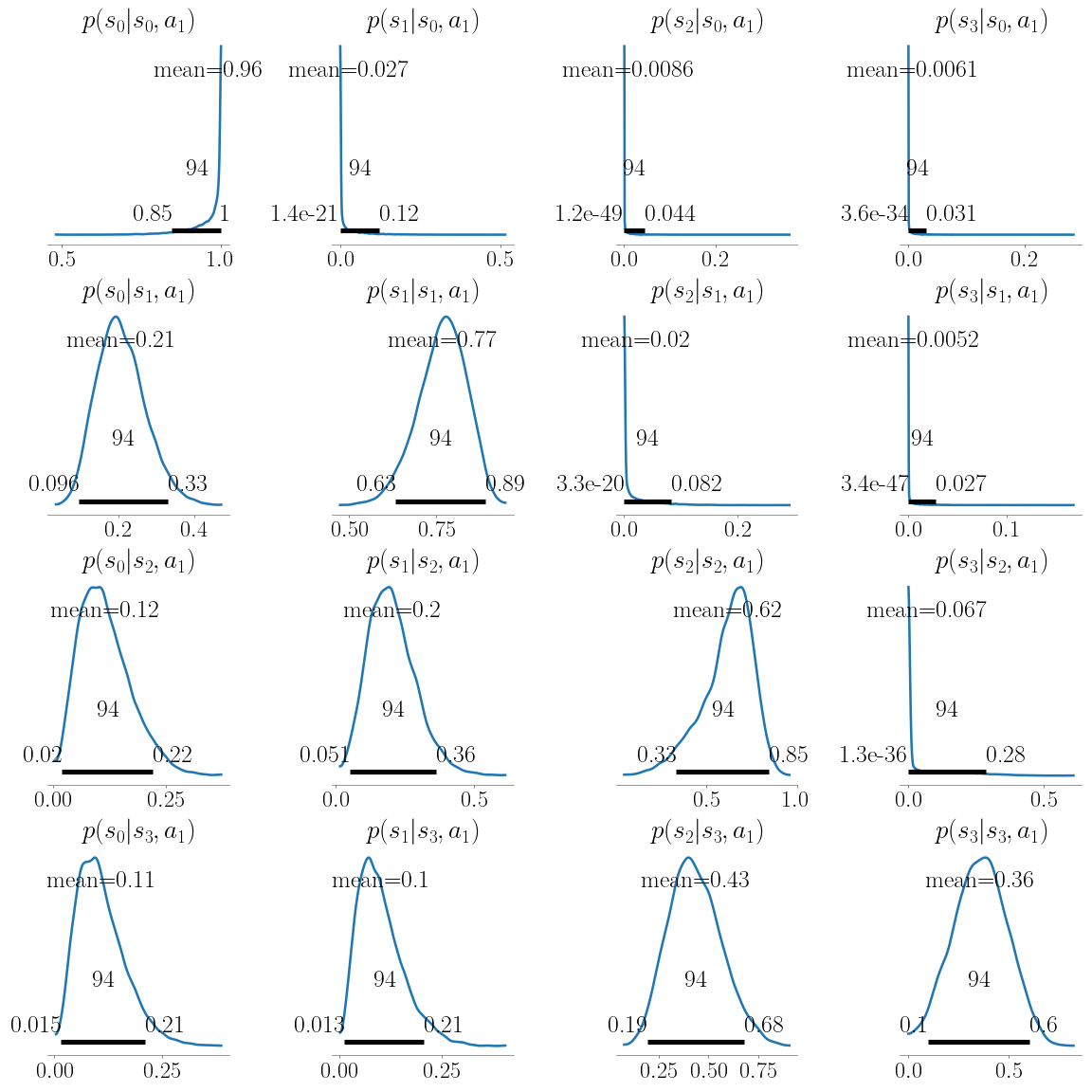}
    \caption{Transition matrix related to action $a_1$ (tamping). The distribution at row $i$ and column $j$ is associated with the probability to transition from state $i$ to $j$ when action $a_1$ is taken. Deterioration of the system (upper right triangle) reflects an almost zero probability, while it appears most
    probable to remain in the same condition or improve by a maximum of one state, which reflects the reduced influence of this action.}
    \label{fig:tr_mat1}
\end{figure}

\newpage

\begin{figure}[!ht]
    \centering
    \includegraphics[width=\linewidth]{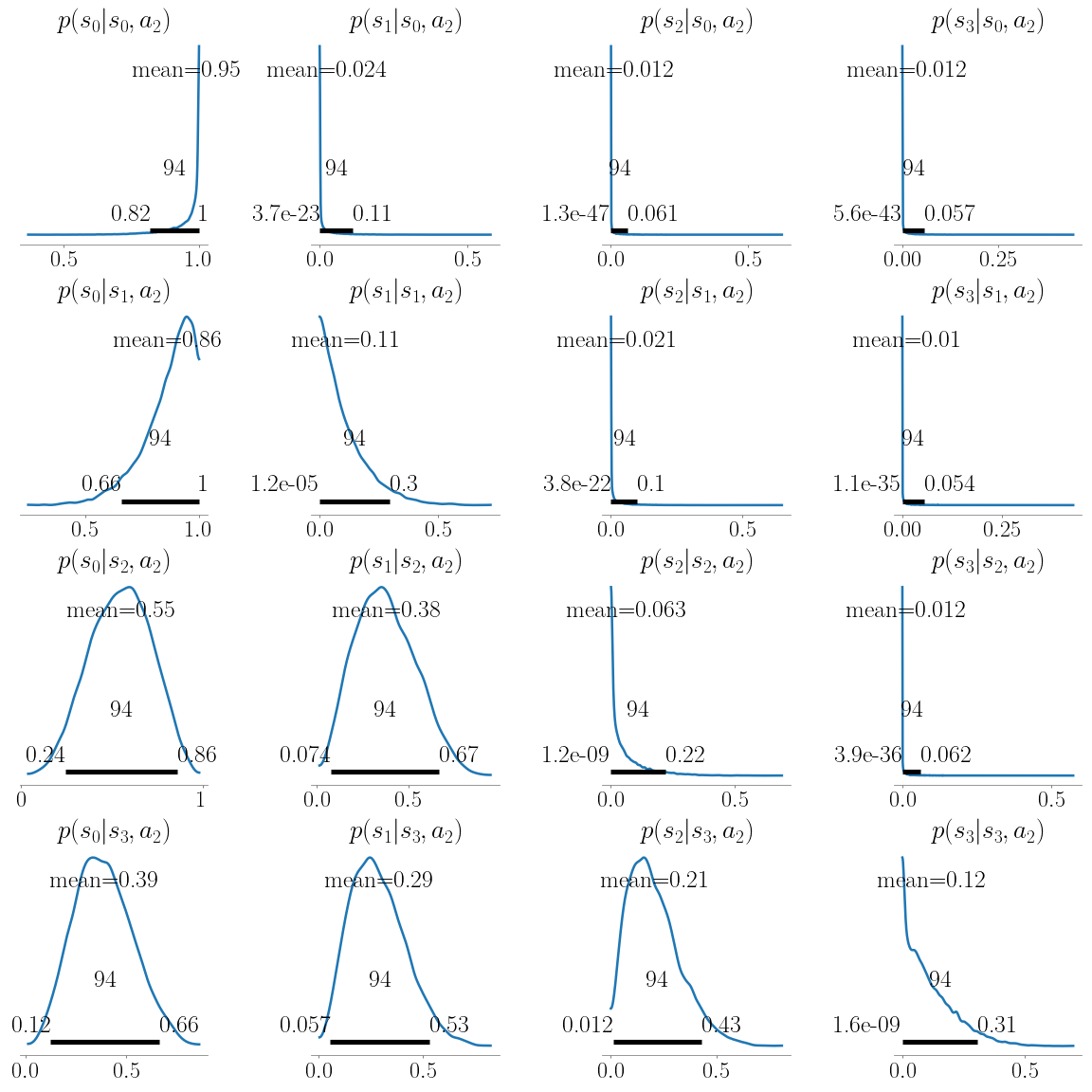}
    \caption{Transition matrix related to action $a_2$ (renewal plus tamping). The distribution at row $i$ and column $j$ is associated with the probability to transition from state $i$ to $j$ when action $a_2$ is taken. Transition to the best possible state $s_0$ is consistently assigned the highest probability, regardless of the starting state, reflecting the higher repairing effect of this maintenance action.}
    \label{fig:tr_mat2}
\end{figure}

\newpage

\subsection{Observation model parameters}
\label{app:inference_observation}

\begin{figure}[!htp]
    \centering
    
\begin{subfigure}[b]{1\textwidth}
\centering
\includegraphics[width=\linewidth]{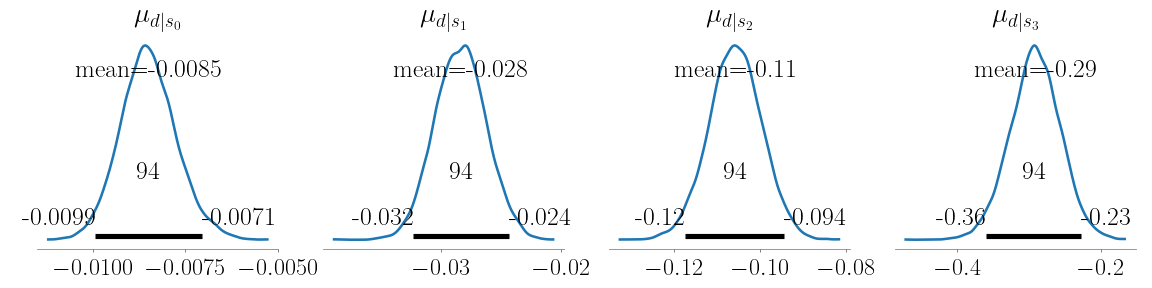}
\caption{Posterior distributions of state-dependent parameters $\mu_{d\mid s_t}$.}
\end{subfigure}
    
\begin{subfigure}[b]{1\textwidth}
\centering
\includegraphics[width=\linewidth]{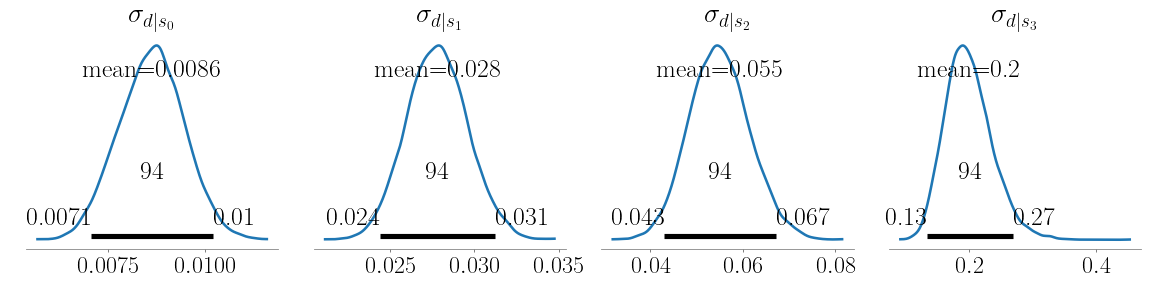}
\caption{Posterior distributions of state-dependent parameters $\sigma_{d\mid s_t}$.}
\end{subfigure}

\begin{subfigure}[b]{1\textwidth}
\centering
\includegraphics[width=\linewidth]{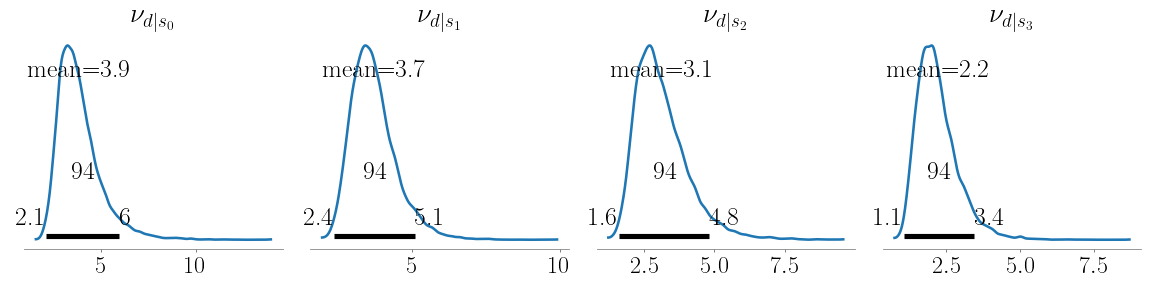}
\caption{Posterior distributions of state-dependent parameters $\nu_{d\mid s_t}$.}
\end{subfigure}

\caption{Posterior distributions of observation model parameters (deterioration process).}
        \label{fig:params_d}

\end{figure}

\newpage

\begin{figure}[!htp]
    \centering

\begin{subfigure}[b]{1\textwidth}
\centering
\includegraphics[width=\linewidth]{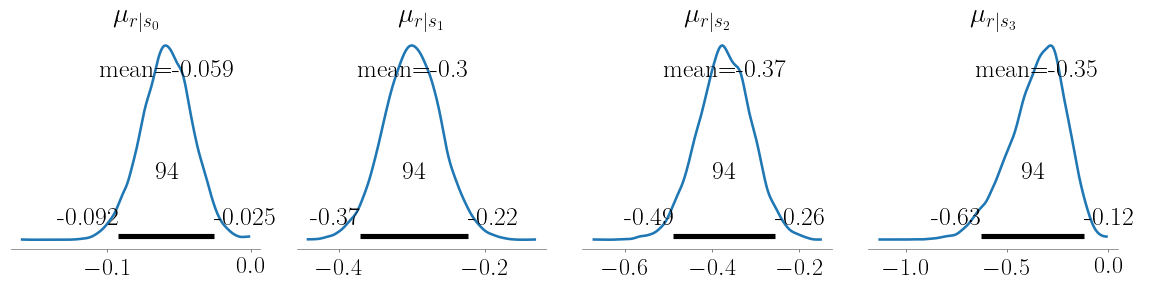}
\caption{Posterior distributions of state-dependent parameters $\mu_{r\mid s_t}$}
\end{subfigure}

\begin{subfigure}[b]{1\textwidth}
\centering
\includegraphics[width=\linewidth]{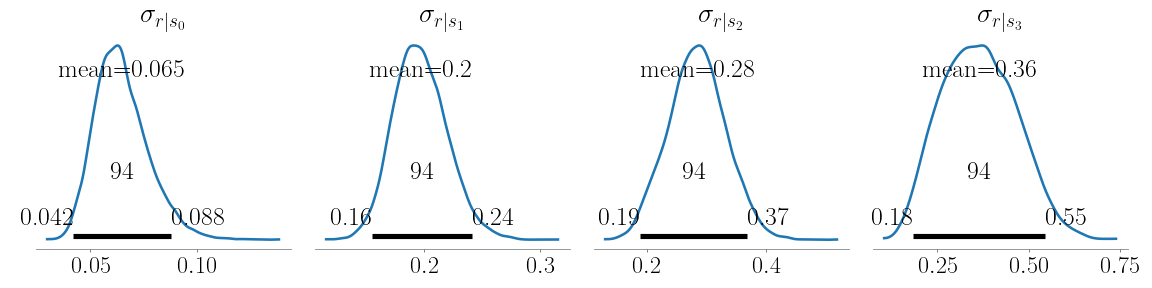}
\caption{Posterior distributions of state-dependent parameters $\sigma_{r\mid s_t}$.}
\end{subfigure}

\begin{subfigure}[b]{1\textwidth}
\centering
\includegraphics[width=\linewidth]{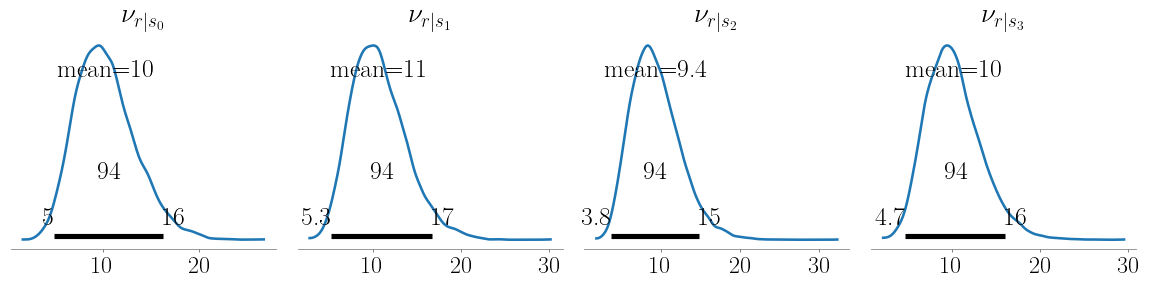}
\caption{Posterior distributions of state-dependent parameters $\nu_{r\mid s_t}$.}
\end{subfigure}

\begin{subfigure}[b]{1\textwidth}
\centering
\includegraphics[width=0.5\linewidth]{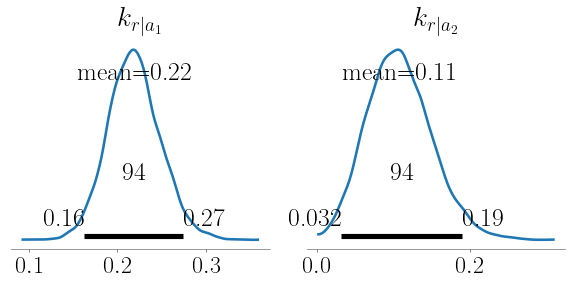}
\caption{Posterior distributions of the autoregressive parameters $k_{r\mid a_t}$ for $a_1$ (left) and $a_2$ (right).}
\end{subfigure}

\caption{Posterior distributions of observation model parameters (repair process).}
\label{fig:params_r}

\end{figure}

\newpage

\begin{figure}[!htp]
    \centering

\begin{subfigure}[b]{1\textwidth}
\centering
\includegraphics[width=\linewidth]{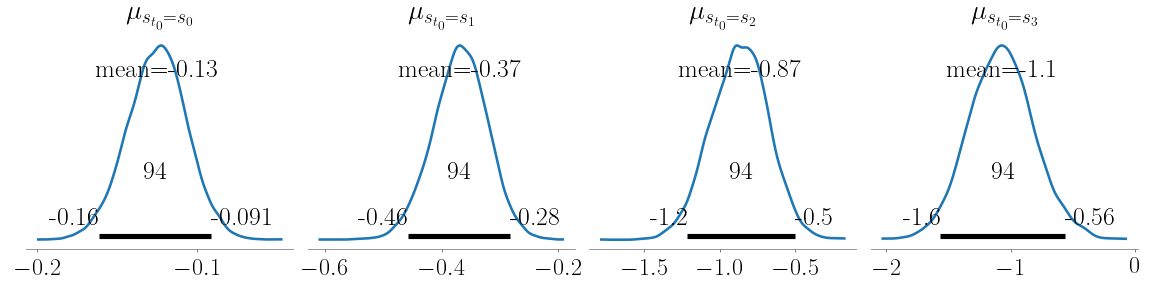}
\caption{Posterior distributions of parameters $\mu_{s_{t_0}}$.}
\end{subfigure}

\begin{subfigure}[b]{1\textwidth}
\centering
\includegraphics[width=\linewidth]{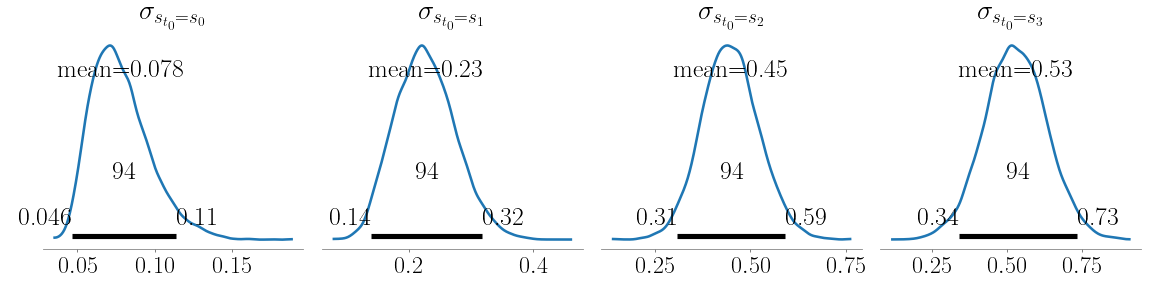}
\caption{Posterior distributions of parameters $\sigma_{s_{t_0}}$.}
\end{subfigure}

\begin{subfigure}[b]{1\textwidth}
\centering
\includegraphics[width=\linewidth]{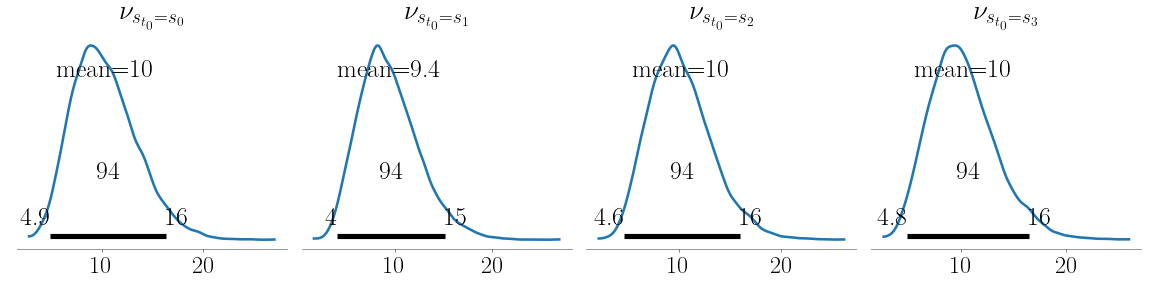}
\caption{Posterior distributions of parameters $\nu_{s_{t_0}}$.}
\end{subfigure}

\caption{Posterior distributions of observation model parameters (initial observation).}
\label{fig:params_i}
\end{figure}

\newpage

\section{Hyperparameters}\label{app:hyper}

\begin{table}[h]
\begin{center}
\caption{Best hyperparameters from the grid-search optimization.}\label{tab:hyper}
\begin{tabular}{@{}lcccc@{}}
\toprule
Hyperparmeter       & Belief (no DR)  & Belief (DR) & GTrXL & LSTM\\
\midrule
Hidden layers        & 3      & 3      & $2\times\textrm{GTrXL}$ & $1\times\textrm{LSTM}+2\times\textrm{MLP}$ \\
Hidden size          &  100   & 100    & -                       & 100 \\
Learning rate        & 0.0001 & 0.0001 & 0.001                   & 0.001 \\
Heads                & -      &  -     & 8                       & - \\
Head dimension       & -      &  -     & 32                      & - \\
Max seq. length      & -      &  -     & 50                      & 3 \\
Memory               & -      &  -     & 50                      & - \\
Use prev. actions    & -      &  -     & Yes                     & Yes \\
Clip parameter       & 0.01   & 0.01   & 0.3                     & 0.3 \\
\botrule
\end{tabular}
\end{center}
\end{table}

\end{appendices}


\bibliography{bibliography}

\end{document}